\documentclass{article}

\usepackage{arxiv}

\usepackage{amsfonts}  
\usepackage{amsmath} 
\usepackage{float}
\usepackage{longtable}
\usepackage{array}
\usepackage{makecell}
\usepackage{siunitx}
\usepackage[T1]{fontenc}    
\usepackage{hyperref}       
\usepackage{url}            
\usepackage{hyperref}
\hypersetup{
    breaklinks=true,
    colorlinks=true,
    linkcolor=blue,
    urlcolor=blue,
    citecolor=blue
}

\usepackage{booktabs}       
\usepackage{amsfonts}       
\usepackage{nicefrac}       
\usepackage{microtype}      
\usepackage{cleveref}       
\usepackage{lipsum}         
\usepackage{graphicx}
\usepackage{natbib}
\usepackage{doi}
\usepackage{parskip} 

\title{A Multimedia Framework for Continuum Robots: Systematic, Computational, and Control Perspectives}

\date{}

\newif\ifuniqueAffiliation
\uniqueAffiliationtrue

\ifuniqueAffiliation 
\author{ \href{https://orcid.org/0009-0001-2682-138X}{\includegraphics[scale=0.06]{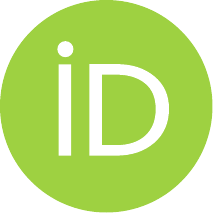}\hspace{1mm}Po-Yu Hsieh} \\
	Graduate Institute of Architecture\\
	National Yang Ming Chiao Tung University\\
	Hsinchu, Taiwan \\
	\texttt{kevinhsieh870118@arch.nycu.edu.tw} \\
	\And
	\href{https://orcid.org/0000-0002-8362-7719}{\includegraphics[scale=0.06]{orcid.pdf}\hspace{1mm}June-Hao Hou} \\
	Graduate Institute of Architecture\\
	National Yang Ming Chiao Tung University\\
	Hsinchu, Taiwan \\
	\texttt{jhou@arch.nycu.edu.tw} \\
}
\renewcommand{\headeright}{}
\renewcommand{\undertitle}{}
\renewcommand{\shorttitle}{A Multimedia Framework for Continuum Robots}

\hypersetup{
pdftitle={A Multimedia Framework for Continuum Robots: Systematic, Computational, and Control Perspectives},
pdfsubject={cs.RO},
pdfauthor={Po-Yu Hsieh, June-Hao Hou},
pdfkeywords={Continuum Robot, Human-Computer Interaction, System Architecture, Dynamics Computation, Control Strategy},
}
\begin{document}
\maketitle

\begin{abstract}
	Continuum robots, which often rely on interdisciplinary and multimedia collaborations, have been increasingly recognized for their potential to revolutionize the field of human-computer interaction (HCI) in varied applications due to their adaptive, responsive, and flexible characteristics. Despite their promises, the lack of an integrated framework poses a significant limitation for both users and developers, resulting in inefficiency and complexity during preliminary developments. Thus, this paper introduces a unified framework for continuum robotic systems that addresses these challenges by integrating system architecture, dynamics computation, and control strategy within a computer-aided design (CAD) platform. The proposed method allows for efficient modeling and quick preview of the robot performance, and thus facilitating iterative design and implementation, with a view to enhancing the quality of robot developments.
\end{abstract}


\section{Introduction}
\begin{figure}[htbp]
	\centerline{\includegraphics[width=.65\linewidth]{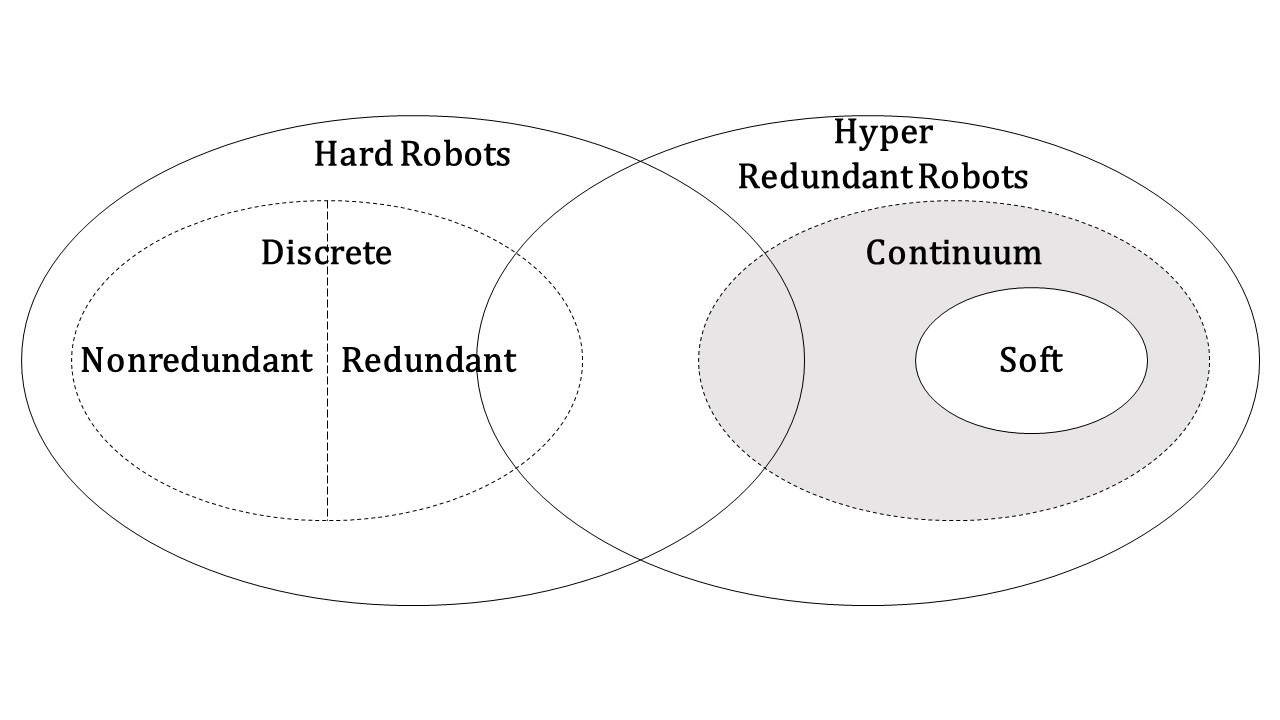}}
	\caption{Classification of robots proposed by \cite{trivedi2008soft}.}
	\label{fig:def}
\end{figure}

Robotics, an interdisciplinary field that merges biology and engineering, aims to develop robotic systems to represent biological entities for assistive purposes. In general, robots can be classified as rigid or soft based on their underlying properties \citep{trivedi2008soft}, as presented in Figure \ref{fig:def}. Rigid robots have been deployed in manufacturing industries due to their accuracy and efficiency within determined spaces. Soft robots, specifically continuum robots, are often used in extreme conditions or confined environments. These robots exhibit enhanced adaptability and high compliance, making them ideal for a range of applications \citep{das2019review}, from medical devices \citep{ashuri2020biomedical} to exploration missions \citep{kobayashi2022soft}. To design and validate the effectiveness of these bio-inspired mechanism, it often requires back and forth testing between computation and prototyping since their inherent elastic nature may cause unexpected deviations or mechanical failure. In fact, an integrated pipeline that allows for efficient modeling and quick preview of the results in both digital and physical environments can significantly enhance the quality of preliminary developments, which is the main focus of this research.

\par The development of such sophisticated systems, however,  have presented technical challenges, particularly due to the lack of an integrated framework that seamlessly synergizes the mechatronic system and robotic design computations. The current landscape of continuum robotics is fragmented, with separate advancements in structural design, computational and mathematical modeling approach \citep{gilbert2021mathematical}, kinematics \citep{orekhov2023lie}, control strategy \citep{yu2024biomimetic}, and mechanical system \citep{fang2023design}. This fragmentation results in a systematic complexity for users with middle-level expertise, raising the overall barrier to entry. Furthermore, current studies often focus more on robot performance and techniques instead of system integration, which may lead to inefficiency and difficulties for independent robot developers to manage a project.

\par Therefore, the principal objective of this research is to develop a unified framework for continuum robots that combines the system architecture, dynamics computation, and control strategy, providing an intuitive, entry-level platform for users and developers.

\section{Prior Work}
\begin{figure}[H]
	\centerline{\includegraphics[width=.5\linewidth]{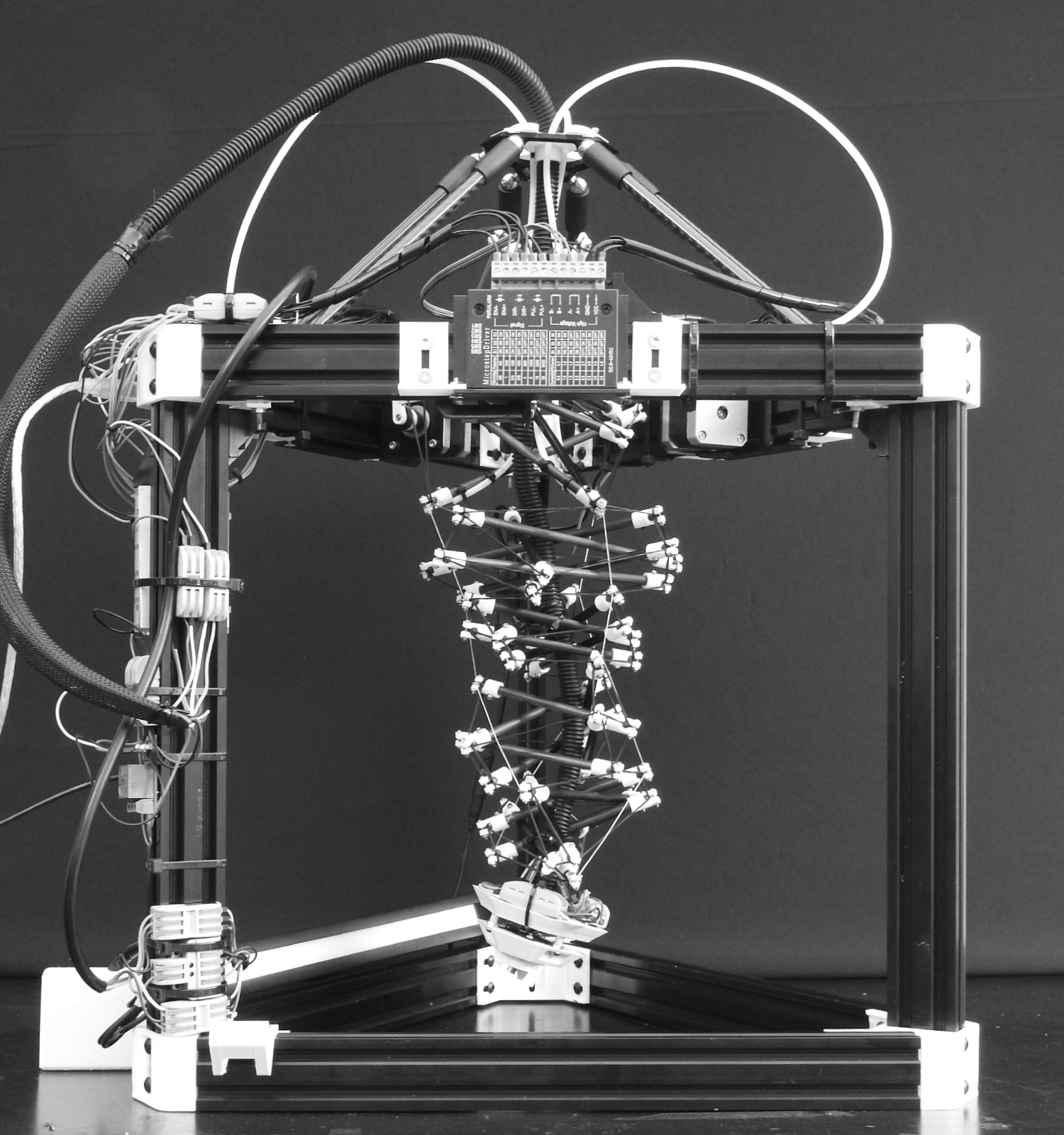}}
	\caption{The previous version of continuum robot proposed by \cite{hsieh2024biomimetic}.}
	\label{fig:priorWork}
\end{figure}

\begin{figure}[H]
	\centerline{\includegraphics[width=.85\linewidth]{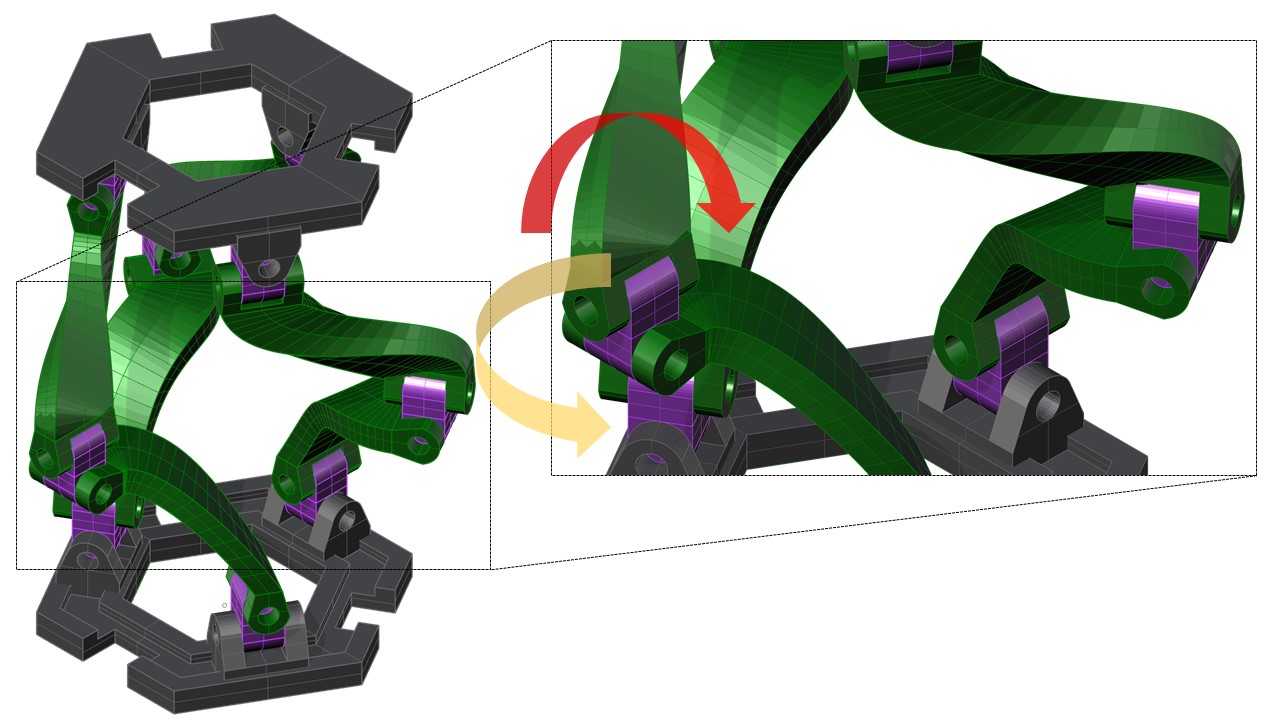}}
	\caption{Detail of the enhanced version.}
	\label{fig:newVersion}
\end{figure}

The robot model used in this research is a tendon-driven tensegrity-based continuum robot, which is an enhanced version of a previous study \citep{hsieh2024biomimetic}. While the previous version (Figure \ref{fig:priorWork}) uses straight rods as the rigid components of the system, the robot in this research (Figure \ref{fig:newVersion}) utilizes two curved objects (green part) interconnected by a pin (purple part) in two different axis. However, the latest version are capable of performing similar movements, indicating the kinematic model in the previous study can still be used in this research. This robot relies on a multi-layer tensegrity structure inspired by spines in vertebrates. It can be manipulated by adjusting each length variation of tendon actuators to achieve various biological movements and adapt to environmental changes. The entire robot can be considered as a dynamic, flexible structure, which shows great technical challenges for design iterations or efficient prototyping using traditional robot operating system.

\section{System Architecture}

\begin{figure}[htbp]
\centerline{\includegraphics[width=\linewidth]{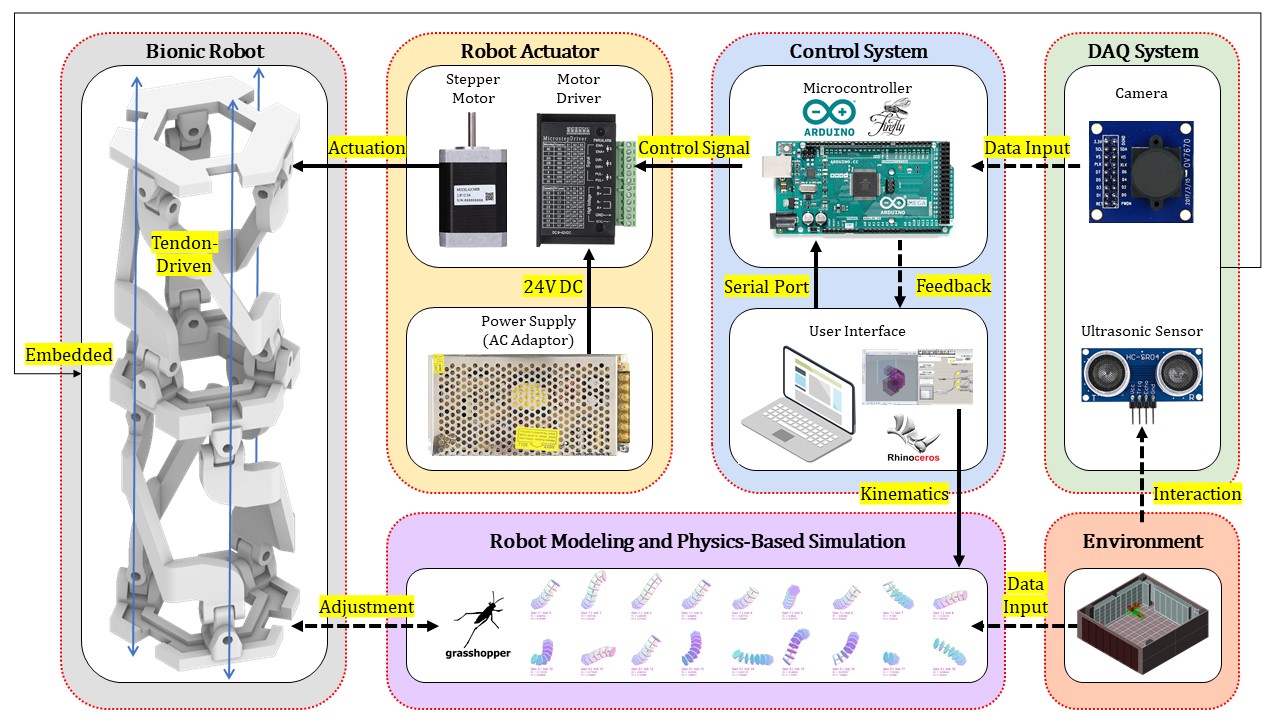}}
\caption{System Architecture.}
\label{fig:systemArchitecture}
\end{figure}

The entire system (Figure \ref{fig:systemArchitecture}) synchronizes the CAD platform Grasshopper in digital (for modeling, physics-based simulations, and control algorithm) and Arduino in physical (for mechatronic, data acquisition, and closed-loop control system) environment since both of them provide accurate, user-friendly, and high-fidelity data for robot development. This intuitive system architecture can be summarized as:

\subsection{Robot Programming (Digital)}
\textbf{Robot Modeling :} The architecture incorporates modeling tools in Grasshopper for designing and simulating the robot's movements. These tools allow for parametric modeling and iterative adjustments, optimizing the robot's design and functionality. \\
\textbf{Physics-based Simulation :} Physics solvers are utilized to predict the robot's behavior under various conditions, refining the control strategies and ensuring robust performance. In this project, we use Kangaroo, which is a physics-based plugin within Grasshopper. 

\subsection{Robot Implementation (Physical)}
\textbf{Control System :} The control system is centralized around an Arduino Mega micro-controller units (MCU), which handles the actions of the stepper motors and processes feedback from the sensors, to ensure precise robotic movements.    \\

\textbf{Robot Actuator :} The actuation is powered by stepper motors (with 3 motor sets in this project), which are regulated by motor drivers. These components convert electrical signals into precise mechanical movements, ensuring accurate positioning and control of the robot's tendons. A 24V DC power supply provides the necessary electrical energy to the stepper motors and motor drivers, ensuring consistent and reliable operation.    \\

\textbf{Data Acquisition (DAQ) system :} The DAQ system includes ultrasonic sensors and cameras, which gather data on the robot's environment and its own positioning. This sensory data is fed back into the control system, enabling the robot to interact effectively with its surroundings. Furthermore, combined with the actuators, this system enables real-time adjustments to the robot's movements, which is crucial for maintaining its accuracy and responsiveness.

\subsection{System Integration}
\textbf{Visual Programming Interface :} To bridge Grasshopper and Arduino, we utilize the Firefly plugin as an efficient integration of the computational and mechatronic control system. By linking the robot through serial ports, data and control signal can be transferred in both directions. This ensures the effectiveness of DAQ system and also allows for validations of the simulated and actual robot performance, which can be collected as datasets for further purposes such as training.

\section{Dynamics Computation}

\begin{figure}[htbp]
\centerline{\includegraphics[width=\linewidth]{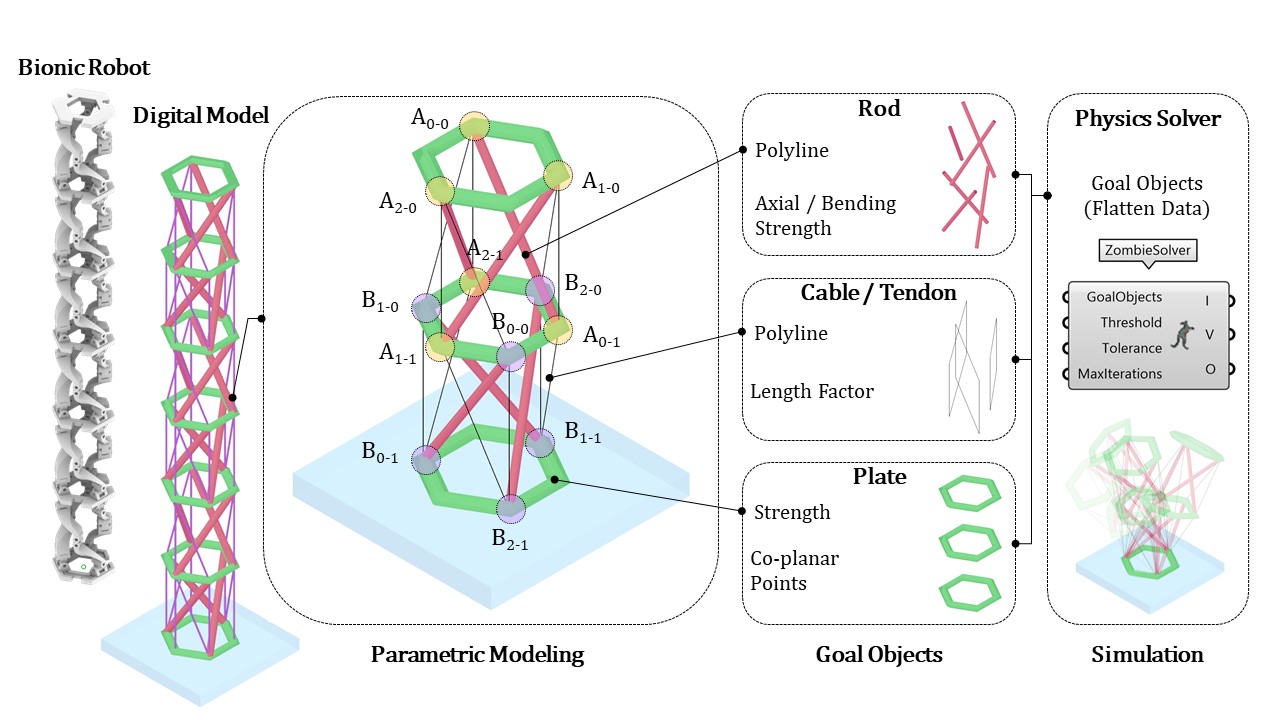}}
\caption{Computational framework (vertices can be referred to Table \ref{table:vertice}).}
\label{fig:computation}
\end{figure}

\begin{figure}[htbp]
\centerline{\includegraphics[width=\linewidth]{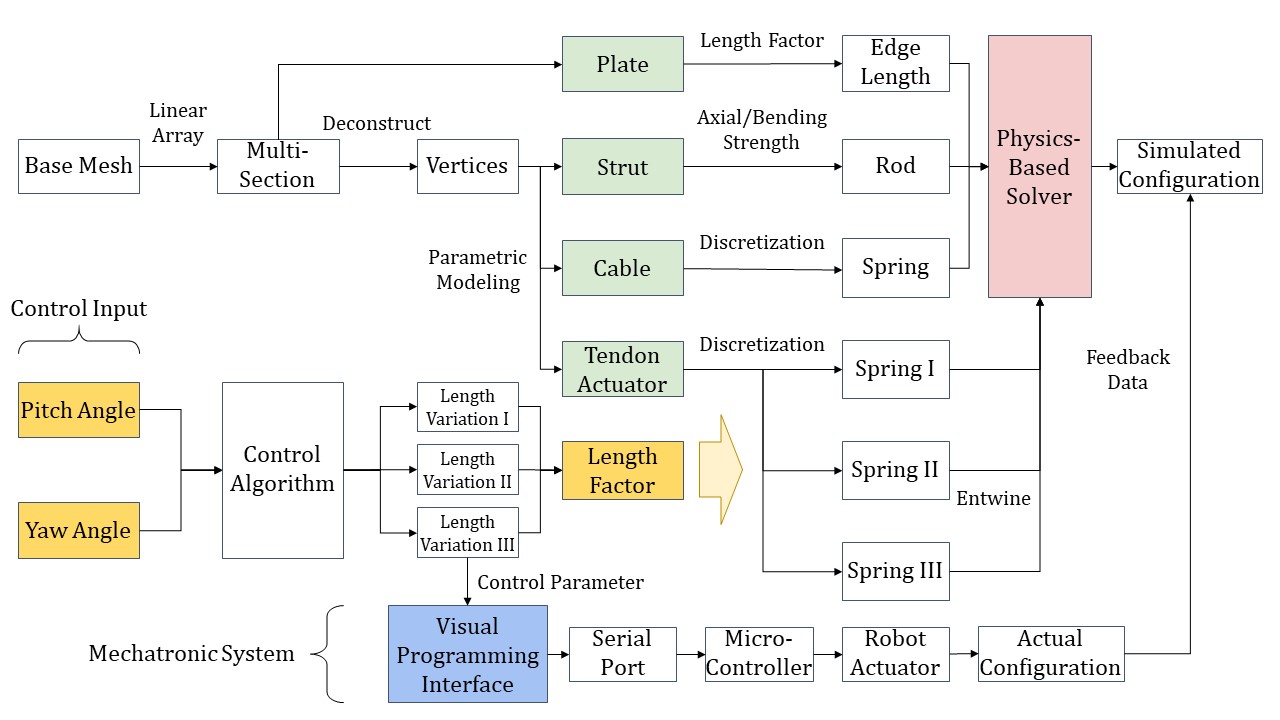}}
\caption{Physics-based simulation in Grasshopper.}
\label{fig:solver}
\end{figure}

One of the most challenging aspect of developing continuum robots is to precisely capture the flexible motions driven by tendon actuators with less computational cost. In this project (Figure \ref{fig:computation}), we use poly-line and 2D mesh objects to represent the tendon actuators and rigid components respectively, which reduces the computation complexity and achieves a responsive modeling method compared to using 3D objects. The overall setup and computation process for physics-based simulation (Figure \ref{fig:solver}) can be described as:
\begin{enumerate}
    \item Defining rigid components (rod and plate): Users can create linear components (rods) equipped with rigidity by setting the axial and bending strength factors. As for the planar rigid parts (plates or segments of the continuum robot), each corner of selected mesh is set as co-planar points to maintain the overall rigidity.
    \item Defining soft components (cable and tendon-actuator): Soft components are linear objects, which connect specific vertices of rigid components following the parametric rules (Table \ref{table:vertice}). By deconstructing these rigid components, the data of mesh vertices can be exported and thus automatically generating linear soft components (poly-lines). Next, by setting the elasticity (length factor) of the selected poly-lines and further discretized each segment, the internal cables and tendon-driven actuators can be precisely created in our digital environment.
    \item  Activating tendon actuators: As rigid and soft components are completed based on parametric modeling, users can control the digital robot model by applying tensile forces and adjusting length variations of tendon actuators. Once the length factor of each actuator is changed, the robot system will reconfigure to achieve a new equilibrium status driven by physics-based solver. Each control input refers to a certain configuration. 
\end{enumerate}

\par By quickly viewing the simulated dynamics and deformation of continuum robots, users can make adjustments to reach their ideal robot configuration or locomotion. Moreover, since the modeling process is entirely based on a parameterized design method (defining geometric rules, vertices information and factors of physical properties), the proposed platform offers a repetitive prototyping workflow. Users can simply test on different parameter sets and understand the impact on robot's performance, and thus making their preferred design decisions.

\renewcommand{\arraystretch}{1.75}

\begin{longtable}{lcccc}
	\caption{Parametric Modeling} \label{table:vertice} \\
	\hline
	\textbf{} & \textbf{Horizontal Cables} & \textbf{Saddle Cables} & \textbf{Vertical Cables} & \textbf{Diagonal Cables} \\ \hline
	\endfirsthead
	\multicolumn{5}{c}%
	{{\bfseries \tablename\ \thetable{} -- continued from previous page}} \\
	\hline
	\textbf{} & \textbf{Horizontal Cables} & \textbf{Saddle Cables} & \textbf{Vertical Cables} & \textbf{Diagonal Cables} \\ \hline
	\endhead
	\hline \multicolumn{5}{c}{{Continued on next page}} \\ \hline
	\endfoot
	\hline
	\endlastfoot
	
	\textbf{Position} & Top, Bottom & Middle & Cross-layer & Cross-layer \\ \hline
	\textbf{Order} & 
	\makecell[l]{\( A_{0-0} \)----\( A_{1-0} \) \\ \( A_{1-0} \)----\( A_{2-0} \) \\ \( A_{2-0} \)----\( A_{0-0} \) \\ \( B_{0-1} \)----\( B_{1-1} \) \\ \( B_{1-1} \)----\( B_{2-1} \) \\ \( B_{2-1} \)----\( B_{0-1} \)} & 
	\makecell[l]{\( A_{0-1} \)----\( B_{0-0} \) \\ \( B_{0-0} \)----\( A_{1-1} \) \\ \( A_{1-1} \)----\( B_{1-0} \) \\ \( B_{1-0} \)----\( A_{2-1} \) \\ \( A_{2-1} \)----\( B_{2-0} \) \\ \( B_{2-0} \)----\( A_{0-1} \)} & 
	\makecell[l]{\( A_{0-0} \)----\( A_{2-1} \) \\ \( A_{1-0} \)----\( A_{0-1} \) \\ \( A_{2-0} \)----\( A_{1-1} \) \\ \( B_{0-0} \)----\( B_{2-1} \) \\ \( B_{1-0} \)----\( B_{0-1} \) \\ \( B_{2-0} \)----\( B_{1-1} \)} & 
	\makecell[l]{\( A_{0-0} \)----\( B_{0-0} \) \\ \( A_{1-0} \)----\( B_{1-0} \) \\ \( A_{2-0} \)----\( B_{2-0} \) \\ \( A_{0-1} \)----\( B_{0-1} \) \\ \( A_{1-1} \)----\( B_{1-1} \) \\ \( A_{2-1} \)----\( B_{2-1} \)} \\ \hline
	\textbf{Amount} & \(h\)=2\(n\) & \(s\)=2\(n\)(\(m\)-2) & \(v\)=\(n\)(\(m\)-1) & \(d\)=\(n\)(\(m\)-1)  \\ \hline
	\textbf{m-Layer} & \multicolumn{4}{c}{\(m\) = 3, 6, 9, \ldots, 3\(p\)+3, where \( p \in \mathbb{N} \)} \\ \hline
	
\end{longtable}
	
\section{Control Strategy}

\begin{figure}[htbp]
\centerline{\includegraphics[width=\linewidth]{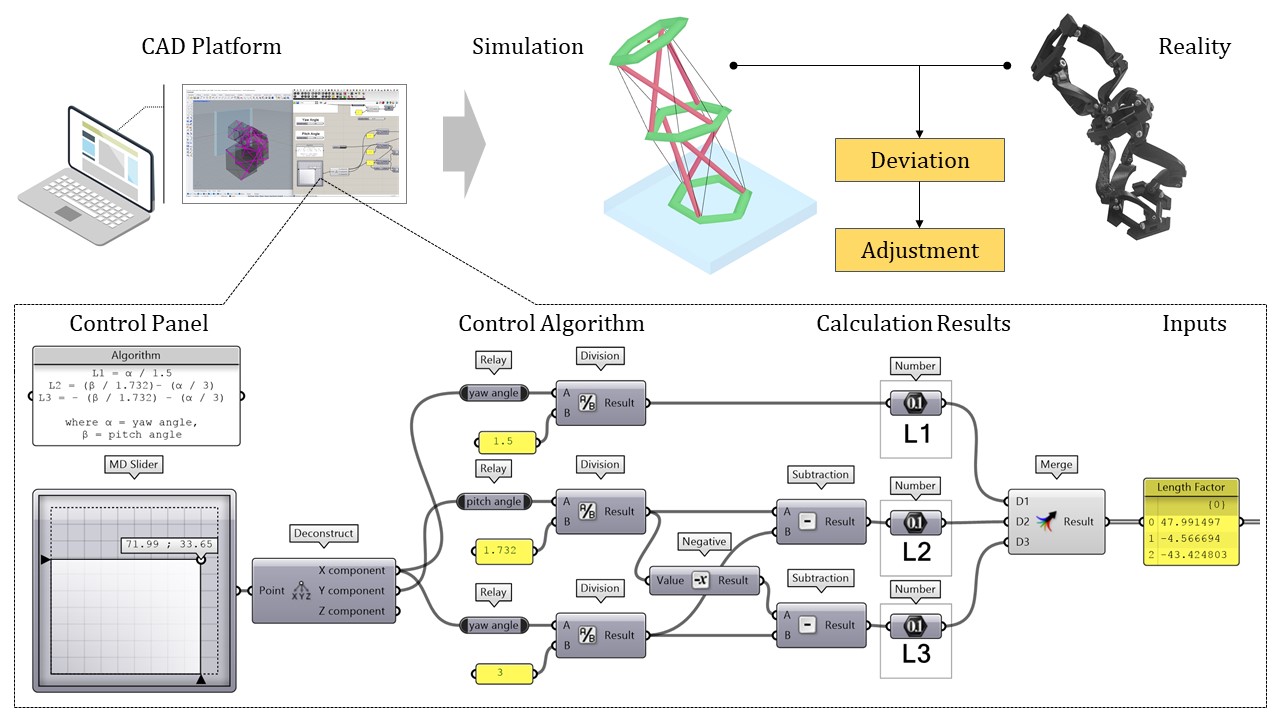}}
\caption{Application of control algorithm.}
\label{fig:controlAlgorithm}
\end{figure}

The versatility and intricacy of continuum robots depend on the performance of different control strategies, which can be considered as the manipulation of length variations of each tendon-driven actuator. To effectively implement various control algorithms, an intuitive interface was included in the proposed framework (Figure \ref{fig:controlAlgorithm}). Developers and users can easily enter different algorithms and simultaneously evaluate the corresponding performance, such as range-of-motions (ROM) and load-bearing capacity. In this project, we  use the fractional order control proposed by \citet{relano2022modeling}, which can produce precise length factors of the actuators by assigning desired yaw and pitch angles ($\alpha$ and $\beta$). Moreover, through comparing the simulated and actual configurations, the coefficients of the algorithm can be further refined to reduce the deviation.
	
\section{Implementation and Discussion}

\begin{figure}[htbp]
\centerline{\includegraphics[width=\linewidth]{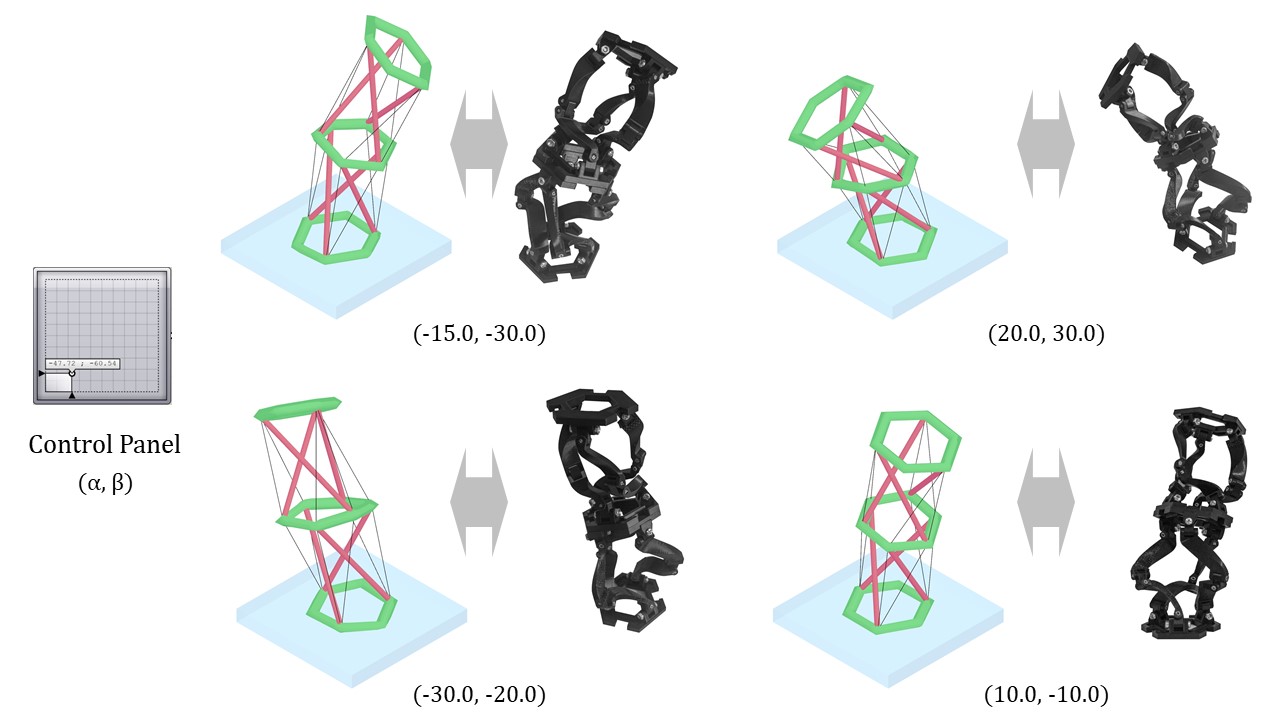}}
\caption{Simulated and actual performance driven by control inputs.}
\label{fig:controlPanel}
\end{figure}

\begin{figure}[htbp]
\centerline{\includegraphics[width=\linewidth]{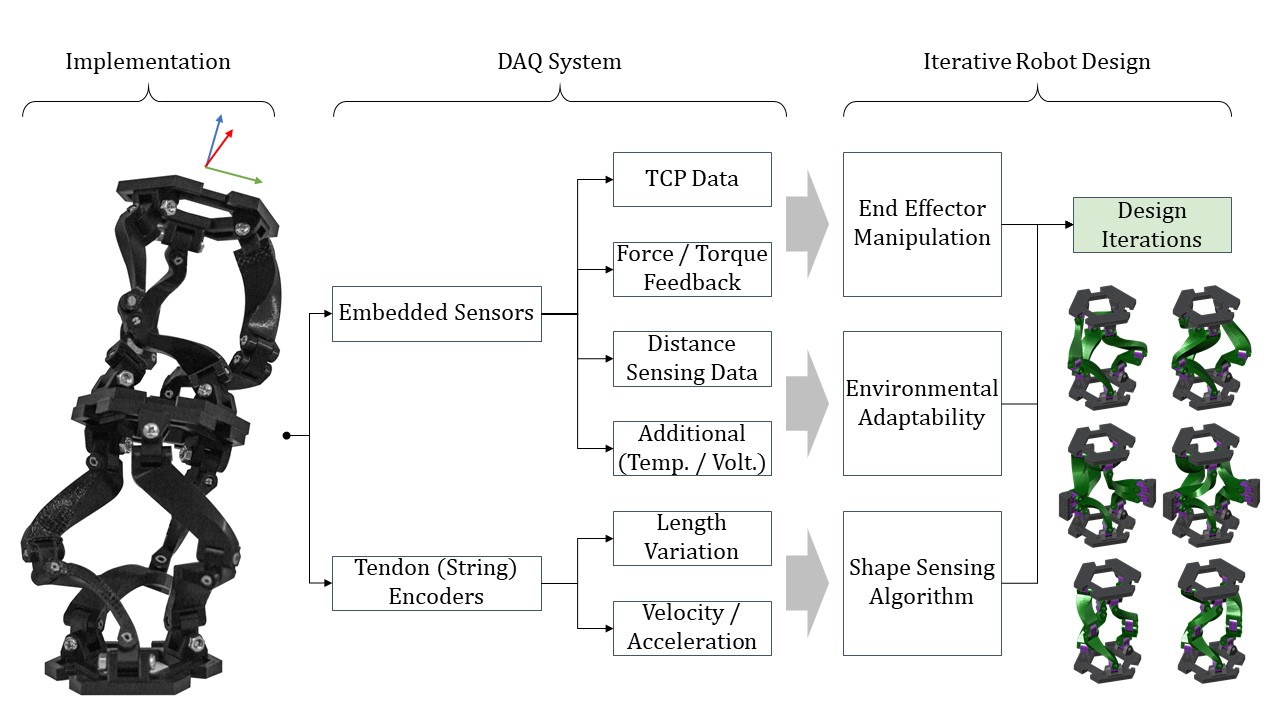}}
\caption{Iterative prototyping process for robot development.}
\label{fig:implementation_iterativeDesign}
\end{figure}

\par In this research, we developed a continuum robot with two segments using the proposed systematic framework as demonstrations. The simulation of the robot's kinematic performance  driven by the physics-based CAD setup can perform, showing the effectiveness of dynamics computation and control system. As presented in Figure \ref{fig:controlPanel}, however, a noticeable deviation from actual robot configuration indicates the fact that the closed-loop control still requires a more accurate adjustment method. Therefore, for further explorations in the future, optimization methods, such as machine learning and evolutionary algorithm, can be integrated into this system to achieve optimal control.

\par
\begin{figure}[htbp]
	\centerline{\includegraphics[width=\linewidth]{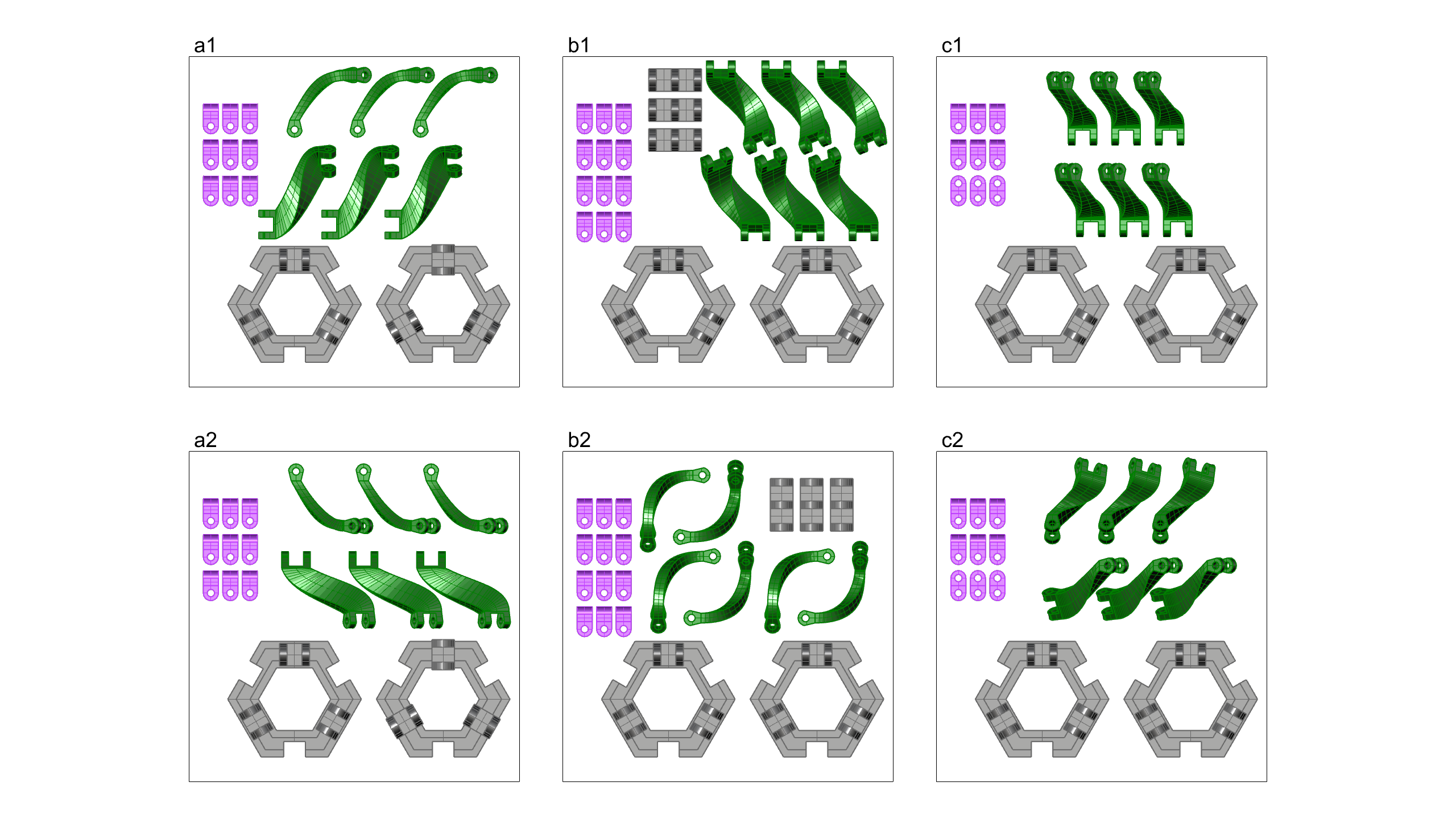}}
	\caption{Fabrication data of design iterations.}
	\label{fig:fabrication}
\end{figure}

 Another significant finding is that the proposed framework can truly benefit the robot development process. As mentioned in the introduction, the efficiency of prototyping process in-between digital and physical models is crucial during preliminary design phase. In this research, a series of design iterations were developed using the proposed framework. As demonstrated in Figure \ref{fig:implementation_iterativeDesign}, by using the DAQ system during the actual robotic operations, the performance can be clearly evaluated based on a variety of sensing data. Therefore, iterative prototypes can be efficiently generated and run through this prototyping process repeatedly. An automated data collection of these design iterations can be considered in future studies to efficaciously record the development and further provide a holistic dataset for generative design tools. On the other hand, this prototyping can also provide fabrication information for efficient robot implementation. As design iterations of digital models are conducted, each corresponding 3D model can be exported and divided into separate parts for 3D-printing, as presented in Figure \ref{fig:fabrication}. Likewise, this feature can be enhanced by adding an automated program, and can even be augmented for mass production in a larger scale.

\section{Conclusions}
Although continuum robots exhibit inherent adaptability and has a profusion of potential for advanced HCI realm, current development of these robots have posed significant challenges due to the fragmented techniques and separate systems. To address these issues, we introduce a comprehensive framework, which integrates both digital and physical environments, and provide demonstrations in this paper. This systematic framework relies on a CAD platform, combining the computational interface and mechatronic system. By synchronizing digital modeling, dynamics computation, and control algorithm, our approach offers an efficient prototyping process and intuitive pipeline for developers to iteratively design, simulate, and validate their specific continuum robots with less complexity.

\par Our integrated framework leverages the computational environment Grasshopper for parametric modeling and physics-based simulations, allowing users to predict and refine the robot's behavior under various conditions. The physical control system, centralized around an Arduino micro-controller, ensures precise and synchronized movements through a visual programming interface and feedback from a data acquisition system. This approach significantly reduces the computational complexity involved in modeling flexible motions of tendon-driven actuators and enables users to intuitively manipulate the robot's movements with a control panel. This real-time motion preview of continuum robots, combined with the simultaneous actuation in reality, provides a responsive method for prototyping in preliminary design phase.

\par Research limitations and guidelines for further research are also discussed in this paper. Based on the robot implementation and its actual performance, there was still a noticeable deviation comparing to the simulated configuration.Thus, more effective optimization methods, such as machine learning and evolutionary algorithms, can be further integrated into the proposed systematic platform to achieve an optimal control. Additionally, automated methods of both data collection and design-to-fabrication can be included in the iterative design workflow for future explorations. Automation in the former can provide detailed training datasets and thus connecting more related techniques, such as generative or AI-aided design tools, while the latter can reduce the time cost for production in a larger amount or scale.

\par Overall, our proposed method not only aims to facilitate more efficient and innovative advancements in the field of continuum robotics but also contributes to lower the barrier to entry for middle-level expertise users and developers in the long run.

\bibliographystyle{unsrtnat}
\bibliography{references}  






\end{document}